\documentclass{article}

\usepackage[final]{corl_2026} 

\usepackage{graphicx} 
\usepackage{acronym}
\usepackage{amssymb}
\usepackage{amsmath}
\usepackage{subcaption}
\usepackage{overpic}
\usepackage{wrapfig}
\usepackage{siunitx}
\usepackage{booktabs}
\usepackage{algorithm}
\usepackage{algpseudocode}
\usepackage{bm}
\usepackage{hyperref}

\newcommand{\update}[1]{\textcolor{black}{#1}}

\title{Learning Loco-Manipulation From SMPC Demonstrations With Sparse Offline-to-Online RL}

\newacro{RL}{Reinforcement Learning}
\newacro{SMPC}{Sample-based Model Predictive Control}
\newacro{TD3}{Twin Delayed DDPG}
\newacro{MPC}{Model Predictive Control}

\author{
  Martin Schuck $^{1,2}$
  \And  
  Maks Sorokin $^{1}$
  \And
  Simone Manni $^{1,3}$
  \And
  Duy Ta $^{1}$
  \And
  Angela P. Schoellig $^{2}$
  \And
  Marco Hutter $^{1,3}$
  \And
  Simon Le Cleac'h $^{1}$
  \And
  Jan Br\"udigam $^{1}$
  \AND
  \normalfont \itshape \small \textsuperscript{1}RAI Institute \quad \textsuperscript{2}Technical University of Munich \quad \textsuperscript{3} ETH Zurich
}

\begin{document}
\maketitle

\vspace{-0.7cm}
\begin{figure}[htbp]
    \centering    \includegraphics[width=0.85\textwidth]{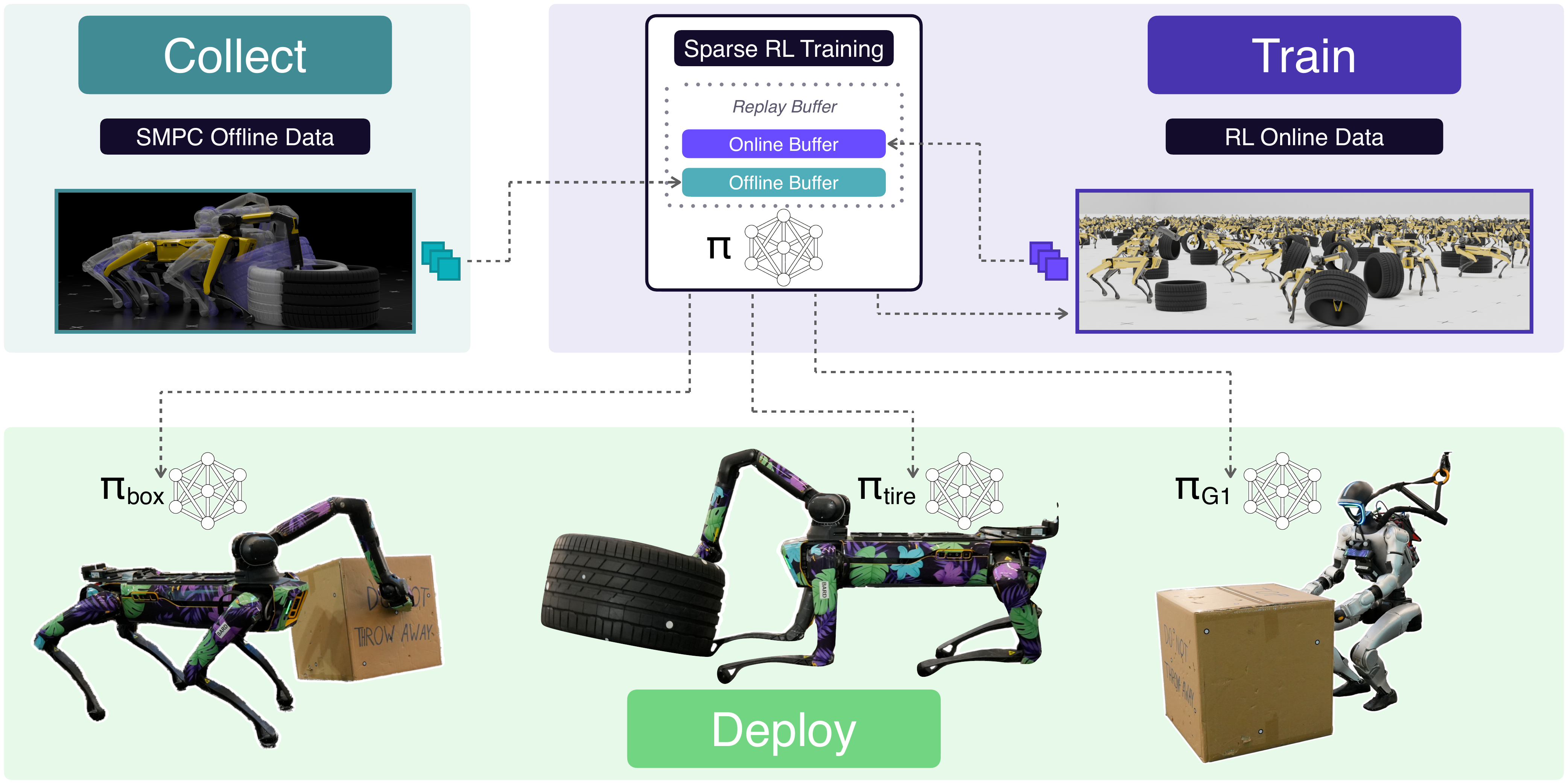}
    \caption{The three main stages of our proposed pipeline. We first collect a dataset from an expert \ac{SMPC}, which is easy to tune in near real-time. We then train complex loco-manipulation policies with sparse rewards solely by mixing in the offline expert data, skipping tedious reward tuning. Finally, the sparse-reward policies are deployed on hardware. Website: \url{https://pages.rai-inst.com/smpc2rl/}}
    \label{fig:placeholder}
\end{figure}


\begin{abstract}
    Integrating locomotion and manipulation is essential for robot autonomy, but scaling standard \ac{RL} to complex tasks is severely bottlenecked by the slow, manual process of dense reward shaping. To bypass this limitation, we leverage \acf{SMPC} entirely in simulation as an automated, rapidly tunable expert to generate massive offline datasets. Because this data solves the fundamental exploration problem, we can train an off-policy \ac{RL} agent using purely sparse task rewards, drastically reducing the time required to learn new skills and eliminating the need for manual tuning. Integrating this high-level agent with a low-level dynamic stability controller yields more optimal behaviors that strictly align with true task objectives, ultimately allowing the learned policies to surpass the original optimal control teacher. We validate the robustness of this sim-to-real framework by successfully deploying complex loco-manipulation skills across different morphologies, including an arm-equipped Spot quadruped and a G1 humanoid.
\end{abstract}

\keywords{RL, Sample-based MPC, Whole-Body Manipulation} 

\section{Introduction}
The ability to seamlessly integrate locomotion and manipulation is essential for deploying autonomous robots in unstructured environments. Yet, learning loco-manipulation policies from scratch using \ac{RL} has proven difficult. Recent developments in \ac{RL} largely focus on imitation and trajectory tracking~\citep{zhao2025resmimic, dao2024sim, liu2025opt2skill}. One major reason for this is that scaling pure \ac{RL} to complex tasks relies heavily on dense reward shaping to guide exploration. Evaluating a single modification to the reward function requires lengthy training cycles, making iterative tuning prohibitively slow. Conversely, optimal control strategies such as \ac{SMPC}~\citep{zhu2025versatile, brudermueller2026generative} are easier to tune, but their high computational cost prevents their direct deployment on high-degree-of-freedom hardware.

Consider, for example, an arm-equipped quadruped tasked with rolling a large tire. Because the object is too massive to drag, the robot must continuously coordinate its stepping to maintain dynamic stability while actively maneuvering its arm to exert forward force. Engineering a dense \ac{RL} reward function for this intricate coordination requires navigating a highly constrained optimization landscape, where each adjustment demands extensive time to validate. In simulation, however, an \ac{SMPC} controller can be iteratively tuned to solve this exact behavior in the order of minutes.

In this paper, we propose leveraging both paradigms to enable training \ac{RL} policies for loco-manipulation. We utilize \ac{SMPC} entirely in simulation as an automated, highly scalable expert to generate massive datasets of successful demonstrations. Because this offline data solves the fundamental exploration problem, we then train an off-policy \ac{RL} agent using purely sparse task rewards. This eliminates the need for manual reward shaping entirely, ensuring the learned policy strictly optimizes the true task objective and naturally \update{achieves faster task completion} than the expert. By doing so, we effectively pair the rapid tuning capabilities of optimal control with the fast execution speeds of neural networks and robustness of \ac{RL} policies trained with domain randomization. To achieve reliable sim-to-real transfer, this high-level agent is integrated with a low-level whole-body controller that actively tracks limb motion while ensuring the robot's dynamic stability~\cite{zhu2025versatile}.

Our contributions are as follows:

\textbf{Tuning-free \ac{RL} training } We introduce a framework that leverages \ac{SMPC} as an \update{easily tunable} automated data generator in simulation, enabling the use of sparse-reward off-policy \ac{RL} and completely circumventing the manual reward-tuning bottlenecks of standard \update{\ac{RL}} pipelines.

\textbf{Optimal loco-manipulation policies } We demonstrate empirically that bootstrapping sparse off-policy \ac{RL} with these \ac{SMPC} datasets yields hardware-deployable policies that ultimately surpass the original optimal control teacher in performance.

\textbf{Validation across real-world platforms } We validate the robustness and generality of our approach across vastly different morphologies, deploying the learned skills on real hardware for both an arm-equipped Spot quadruped and a G1 humanoid.


\section{Related Work} \label{sec:related_work}
\ac{MPC} approaches leverage known system dynamics to achieve loco-manipulation tasks such as pulling and pushing \citep{molnar2025whole}, grasping and picking-and-placing \citep{sambhus2026nonlinear, rigo2024hierarchical}, or non-prehensile object moving \citep{rigo2022contact}. However, by definition, these methods require accurate model knowledge, including the dynamics' gradients, and are limited to prespecified contacts or narrow task formulations due to their high computational demand. In contrast, \ac{RL}-based methods offer fast execution of advanced tasks by distilling optimization into networks. As a result, policies can perform more general manipulation of objects \citep{dao2024sim, liu2025opt2skill}, including whole-body interactions \citep{zhao2025resmimic}. Yet, since on-policy \ac{RL} for complex loco-manipulation requires dense reward shaping to guide exploration, current approaches use \ac{RL} to track provided interaction trajectories instead of learning behaviors purely from rewards. As such, their generality is limited to the quality and availability of demonstrations. Our approach leverages the advantages of both paradigms: rather than deploying model-based optimal control in real-time, we use it entirely in simulation as an automated offline teacher to bypass the \ac{RL} reward-tuning bottleneck, while allowing \ac{RL} to outperform the expert demonstrations.

Leveraging large offline datasets for complex robotics requires architectures optimized for rapid learning and robust sim-to-real transfer. The offline-to-online paradigm~\cite{ball2023efficient} has proven effective at accelerating the training of \ac{RL} agents with expert demonstrations. Streamlined off-policy architectures such as FastTD3 and FastSAC~\cite{seo2025fasttd3} have demonstrated strong sample efficiency, enabling training and sim-to-real transfer of high-dimensional humanoid locomotion with significantly reduced training times. We extend these approaches by integrating the offline-to-online paradigm with a modified FastTD3 architecture. By bootstrapping this efficient learner with our \ac{SMPC}-generated expert datasets, we enable the acquisition of complex loco-manipulation skills using purely sparse rewards. This approach significantly reduces the overall training time and eliminates the need for manual \ac{RL} reward engineering.

There exists other work that circumvents the exploration challenges and manual reward shaping inherent in standard \ac{RL} for manipulation by using demonstrations to guide the learning process. Learning from human demonstrations can be achieved with an augmented behavior-cloning loss~\cite{nair2018overcoming}, by bootstrapping a policy with imitation learning~\cite{hu2023imitation}, or by adding demonstrations to the replay buffer~\cite{vecerik2017leveraging}. However, the tasks investigated in these works are tabletop end-effector manipulation tasks---obtaining human demonstrations for loco-manipulation tasks, especially on non-humanoid robots, is a restrictive limitation for these methods. An alternative is using algorithmic planners to generate demonstrations. Learning can be guided by such demonstrations through informed reset states~\cite{pinto2018sample, khandate2024r}. As before, these methods are limited to tabletop and in-hand manipulation tasks. In~\cite{bruedigam2024jacta}, a graph-based motion planner with task-specific heuristics is used to generate dexterous manipulation data for training \ac{RL} policies in stationary manipulation settings. Our approach focuses on dynamic, whole-body loco-manipulation rather than stationary dexterity. Furthermore, instead of relying on tree search and discrete heuristics, we utilize a unified \ac{SMPC} framework that readily scales to a wide variety of loco-manipulation tasks and embodiments.


\section{Learning Loco-Manipulation With SMPC Demonstrations} \label{sec:method}

\begin{figure}
    \centering
    \begin{subfigure}[b]{0.32\textwidth}
        \centering
        \includegraphics[width=\textwidth]{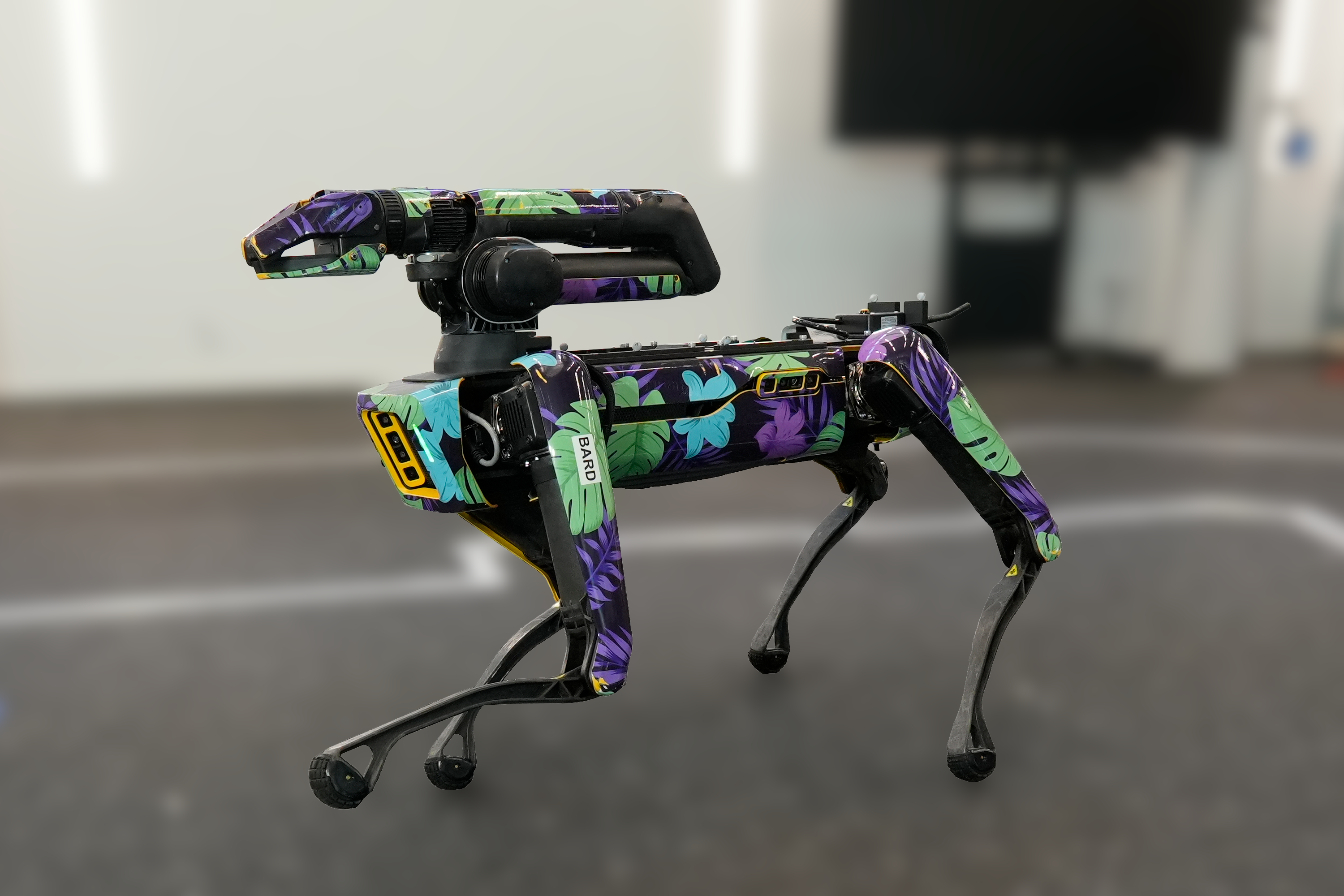}
    \end{subfigure}
    \hfill
    \begin{subfigure}[b]{0.32\textwidth}
        \centering
        \includegraphics[width=\textwidth]{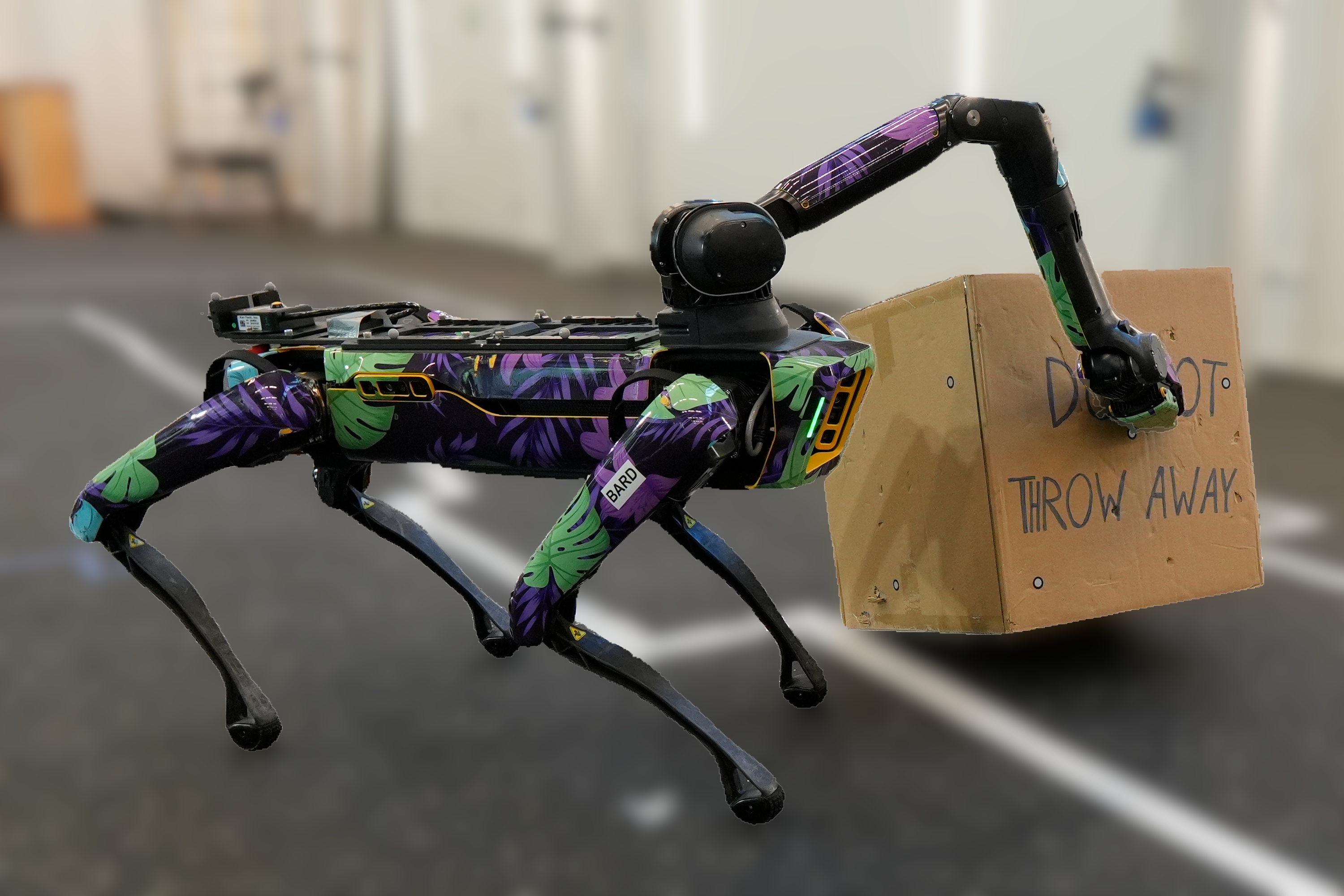}
    \end{subfigure}
    \hfill
    \begin{subfigure}[b]{0.32\textwidth}
        \centering
        \includegraphics[width=\textwidth]{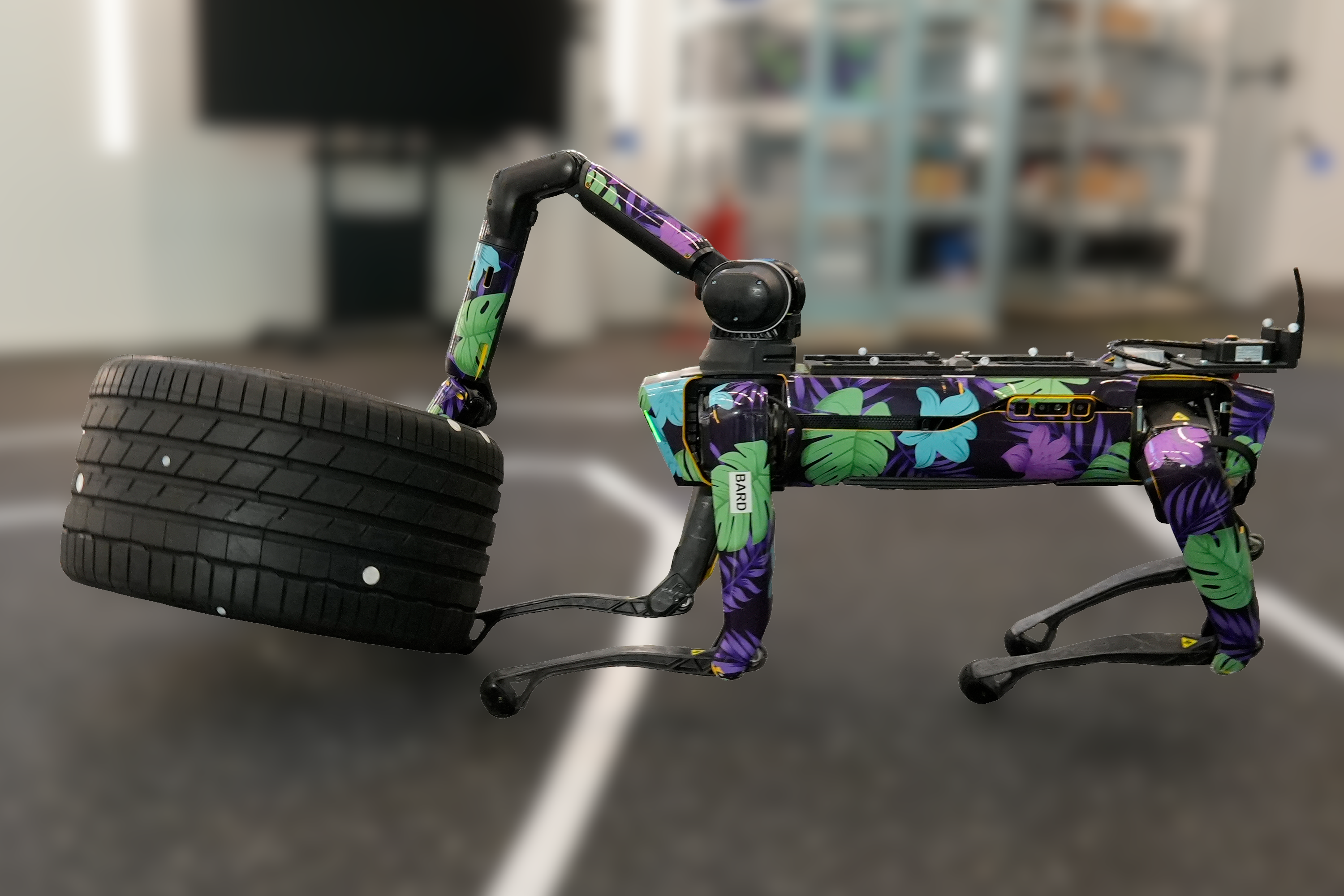}
    \end{subfigure}
    \\
    \vspace{4pt}
    \begin{subfigure}[b]{0.49\textwidth}
        \centering
        \includegraphics[width=\textwidth]{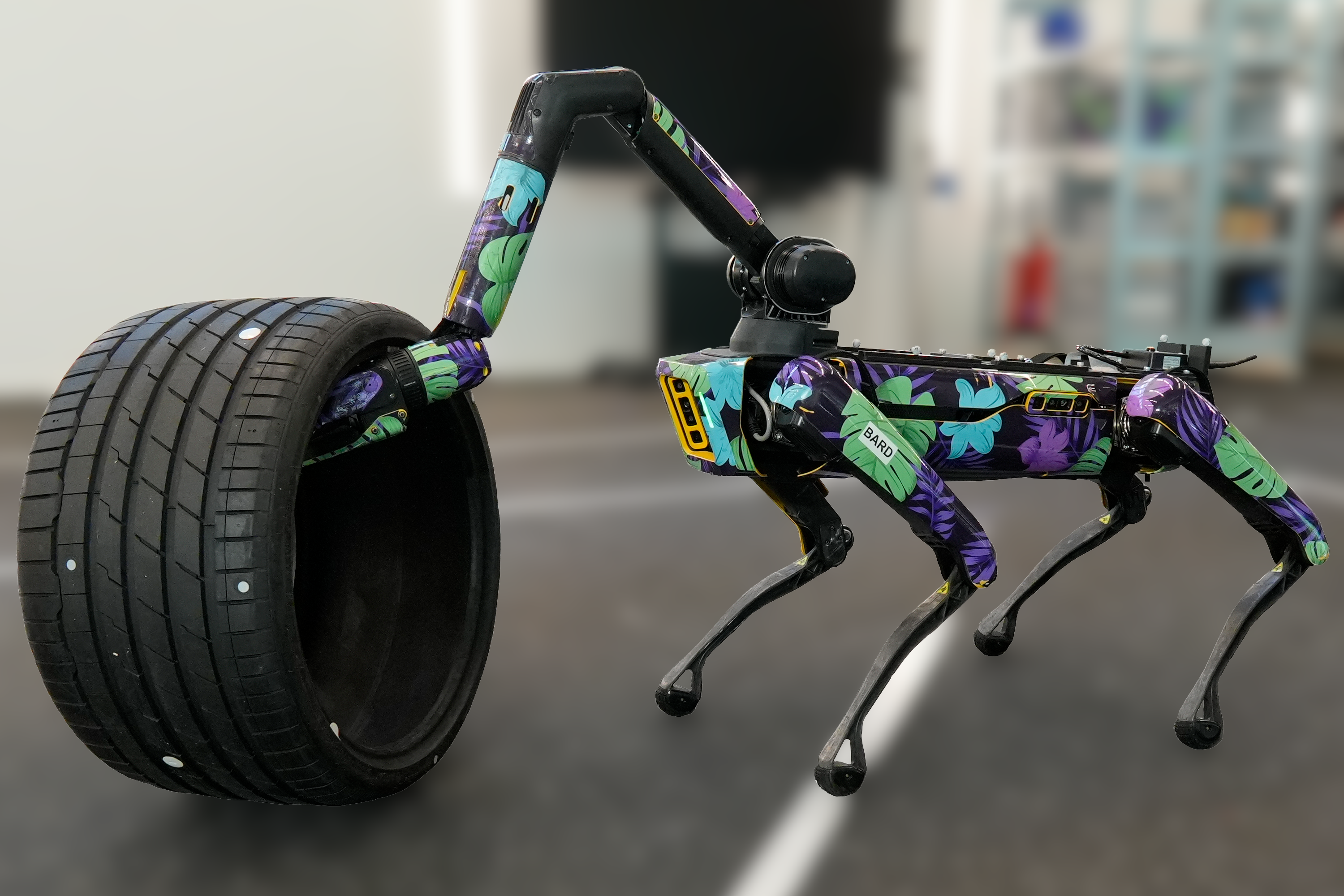}
    \end{subfigure}
    \hfill
    \begin{subfigure}[b]{0.49\textwidth}
        \centering
        \includegraphics[width=\textwidth]{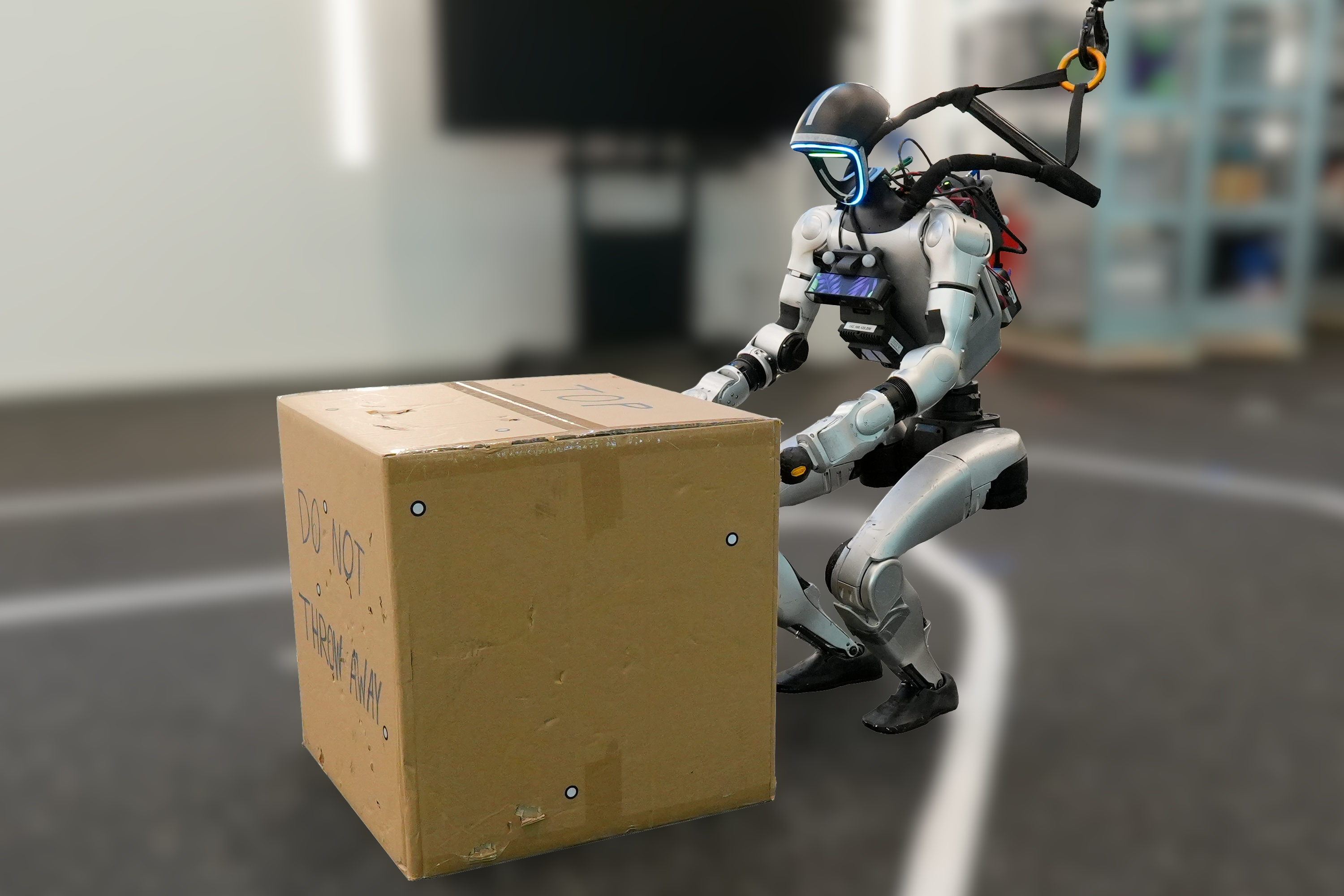}
    \end{subfigure}

    \caption{Real-world deployment of our tasks. Our framework learns complex loco-manipulation skills across different embodiments (Spot and G1), objects (boxes and tires), and skills (pushing, rolling, or lifting). From left to right, top to bottom: Spot reach, box pushing, tire uprighting, tire rolling, and G1 box pushing.}
    \label{fig:tasks}
    \vspace{-10pt}
\end{figure}

Our objective is to learn robust loco-manipulation policies using strictly sparse task rewards, thereby eliminating the need for extensive, manual reward tuning when acquiring new skills. However, training \ac{RL} agents with sparse objectives poses a severe exploration challenge, particularly in the high-dimensional continuous action spaces of systems that combine legs and arms. To overcome this, we adopt an offline-to-online training paradigm, utilizing off-policy algorithms bootstrapped with expert demonstrations to bypass the initial exploration phase. Rather than relying on human teleoperation or human motion retargeting—which scale poorly to non-anthropomorphic platforms such as arm-equipped quadrupeds—we construct our offline dataset using \ac{SMPC} in simulation. \ac{SMPC} allows for rapid, intuitive tuning of complex behaviors and can be straightforwardly applied to any robotic morphology.

The remainder of this section is structured as follows: First, we provide an overview of our hierarchical control architecture (\autoref{sec:method:overview}). Next, we formulate our offline-to-online \ac{RL} pipeline using \ac{TD3}~\citep{fujimoto2018addressing} to train policies exclusively on sparse-rewards (\autoref{sec:method:rl}). Finally, we demonstrate how we leverage \ac{SMPC} to autonomously generate the required high-quality datasets across diverse morphologies (\autoref{sec:method:smpc}).

\subsection{Overview} \label{sec:method:overview}
We employ a hierarchical control architecture that decouples task-level navigation and manipulation from base maneuvering. This structure is shared during data collection and deployment, allowing seamlessly swapping between the \ac{SMPC} expert or the \ac{RL} agent for the high-level policy.

\textbf{High-Level Task Policy} The high-level policy is responsible for overarching manipulation and navigation. Its action space is defined as $a_{\mathrm{high}} = [\Delta v_{\mathrm{cmd}}, \Delta q_{\mathrm{cmd}}^{\mathrm{arm}}, \Delta h_{\mathrm{cmd}}, \Delta p_{\mathrm{cmd}}]$, where $\Delta v_{\mathrm{cmd}} \in \mathbb{R}^3$ represents the delta to the current desired planar base velocity, $\Delta q_{\mathrm{cmd}}^{\mathrm{arm}} \in \mathbb{R}^{n_{\mathrm{a}}}$ specifies the delta to the current target joint positions for the $n_{\mathrm{a}}$-degree-of-freedom arm, $\Delta h_{\mathrm{cmd}} \in \mathbb{R}$ specifies the delta to the current torso height target, and $\Delta p_{\mathrm{cmd}} \in \mathbb{R}$ specifies the delta to the current desired torso pitch target. For Spot, we keep the torso height and pitch fixed and do not command these values.

\textbf{Low-Level Stabilization Controller} The kinematic commands from the high-level policy are passed to a whole-body maneuvering controller trained via the ReLIC framework~\citep{zhu2025versatile}, which is subsequently frozen for all downstream task learning. Given $a_{\mathrm{high}}$ and the current proprioceptive state, it outputs full-body joint position targets $q_{\mathrm{target}} = [q_{\mathrm{target}}^{\mathrm{legs}}, q_{\mathrm{target}}^{\mathrm{arm}}]$. This low-level policy strictly tracks the commanded arm motion and base velocity while dynamically adjusting the leg joints $q_{\mathrm{legs}}$ to maintain balance, enabling the high-level policy to focus exclusively on task completion.

\subsection{Training Off-Policy RL with Sparse Rewards} \label{sec:method:rl}

We formulate the high-level loco-manipulation task as a Markov decision process and train our policy in simulation using MuJoCo Warp and mjlab~\citep{todorov2012mujoco, zakka2026mjlab}. Because our objective is to eliminate manual reward engineering, we define a strictly sparse task reward function
\begin{equation}
    r = \begin{cases}
        0 \quad &\text{at goal}\\
        -\frac{2}{1-\gamma} \quad &\text{robot crashed}\\
        -1 & \text{else}
    \end{cases},
\end{equation}
where $\gamma$ is the discount factor of $0.99$ in our experiments. The crash penalty for terminations is chosen such that agents prefer the constant negative reward over crashing, i.e., $\sum_{i=0}^\infty -1\gamma^i = -\frac{1}{1-\gamma}$. \update{The robot is considered crashed when the torso height or tilt exceeds pre-defined limits.}

Naturally, optimizing this sparse reward from scratch in a high-dimensional continuous action space is intractable without tailored exploration strategies. To overcome this bottleneck, we employ an offline-to-online training strategy utilizing a modified FastTD3 architecture. During the initial phases of learning, we replace 50\% of the transitions in the replay buffer with pre-collected expert data (detailed in \autoref{sec:method:smpc}). This steady injection of expert transitions provides the necessary successful samples to ground the critic networks, allowing them to produce meaningful policy gradient estimates before the online agent randomly discovers successful trajectories. To prevent the policy from continually relying on these demonstrations, we use a curriculum that phases out the expert data once the agent achieves a sufficient empirical success rate, shifting to pure online learning.

Because policies trained with sparse objectives seek optimal solutions that push the physical limits of the platform, we must ensure these behaviors remain safe and deployable. To achieve this, we parameterize the high-level policy to output action deltas ($[\Delta v_{\mathrm{cmd}}, \Delta q_{\mathrm{cmd}}^{\mathrm{arm}}, \Delta h_{\mathrm{cmd}}, \Delta p_{\mathrm{cmd}}]$) relative to the current desired velocity and kinematic state, rather than absolute targets. This parameterization allows us to explicitly enforce maximum desired accelerations and speeds within the action space.

Finally, because our sparse reward formulation defines strictly known maximum and minimum possible returns, we structurally bound the critic network outputs to these limits. Unbounded overestimation in Q-functions is strongly associated with learning instability~\citep{Hasselt2018DeepRL}, and bounded critics have been successfully applied in advanced algorithms~\citep{bhatt2024crossq}. Empirically, we observe that enforcing these bounds at the architectural level significantly stabilizes training, particularly when hyperparameter configurations are still suboptimal. Complete environmental details, network architectures, and hyperparameters are provided in \autoref{app:rl}.

\subsection{Data Collection with Sampling-Based Model Predictive Control} \label{sec:method:smpc}

\begin{figure}
    \centering
    \includegraphics[trim=4.0cm 2cm 10cm 2.5cm, clip, width=\textwidth]{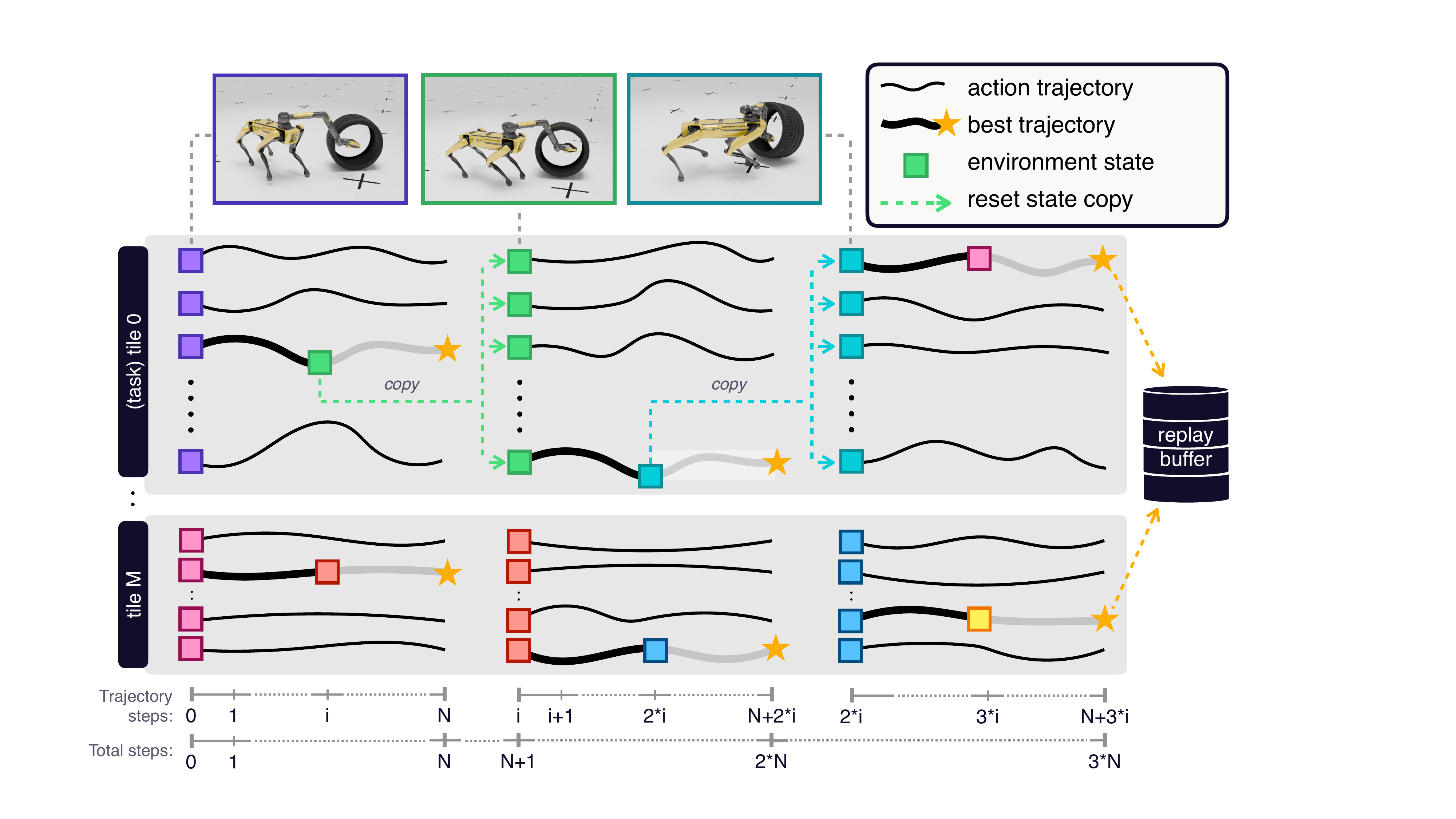}
    \caption{We scale our \ac{SMPC} data collection by solving multiple \acp{SMPC} in parallel on the GPU using vectorized environments divided into tiles. Each tile broadcasts its initial task to all its environments upon reset, runs sampled actions, selects the best trajectories, and adds them to our expert buffer. Note the different x-axes scales for a task's trajectory steps and total simulation steps.}
    \label{fig:smpc}
    \vspace{-10pt}
\end{figure}

To collect large-scale expert demonstrations without the prohibitive expense of manual teleoperation, which scales poorly to complex morphologies like an arm-equipped Spot, we employ \ac{SMPC} as an automated expert. We run the \ac{SMPC} using the same vectorized MuJoCo Warp environments as our \ac{RL} pipeline, though it is guided by easily formulated dense cost functions rather than sparse rewards. At first glance, this might seem counterintuitive because we are reintroducing dense rewards that need to be tuned. However, the key difference is that the \ac{SMPC} can be tuned in near real time, does not require training, and can thus be run interactively. On a single RTX 5090, our \ac{SMPC} achieves $\sim0.5\times$ real-time performance. This allows us to tune rewards in minutes and start generating expert data without training a policy.

We find that a streamlined \ac{SMPC} formulation, similar to \citet{howell2022predictive}, with random sampling and a warm-start strategy typical of \ac{MPC} algorithms, is sufficient to solve our tasks without additional complexities such as Model Predictive Path Integral Control~\citep{Williams2017ModelPP}. Once the dense costs are tuned, we scale the data collection using a massively parallelized tiled approach (see \autoref{fig:smpc}). This architecture allows us to generate high-quality demonstration datasets at a rate of one million samples per hour (where each sample consists of a single observation, action, reward, and next observation), providing the offline data required to successfully bootstrap the sparse-reward \ac{RL} agent within four hours on a single GPU. Complete algorithmic details are provided in \autoref{app:smpc}.

\section{Experiments} \label{sec:experiments}
We demonstrate the viability of our approach on a Boston Dynamics Spot quadruped equipped with an arm and a Unitree G1 humanoid. Additionally, we answer the following research questions with structured ablations:

\begin{figure}
    \centering
    \begin{subfigure}[b]{0.32\textwidth}
        \centering
        \includegraphics[width=\textwidth]{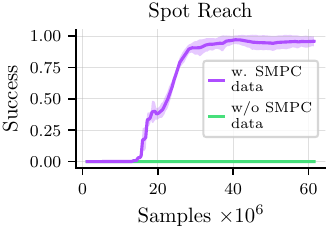}
    \end{subfigure}
    \hfill
    \begin{subfigure}[b]{0.32\textwidth}
        \centering
        \includegraphics[width=\textwidth]{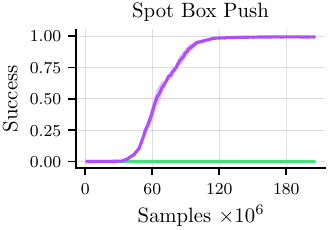}
    \end{subfigure}
    \hfill
    \begin{subfigure}[b]{0.32\textwidth}
        \centering
        \includegraphics[width=\textwidth]{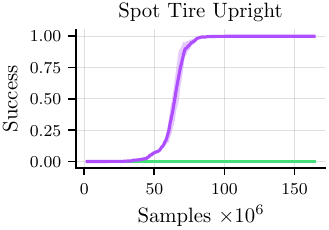}
    \end{subfigure}
    \\
    \begin{subfigure}[b]{0.33\textwidth}
        \centering
        \includegraphics[width=\textwidth]{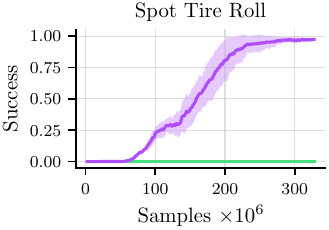}
    \end{subfigure}
    \begin{subfigure}[b]{0.33\textwidth}
        \centering
        \includegraphics[width=\textwidth]{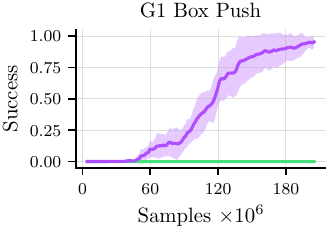}
    \end{subfigure}
    \caption{Training performance of our framework on different tasks. We show the success rate of our policy over time in increasingly complex tasks. Our framework reliably reaches success rates close to 100\% in simulation without any reward shaping. Without adding expert \ac{SMPC} data, learning complex loco-manipulation policies from sparse rewards fails to learn anything. Statistics are averaged over five seeds, with the shaded area representing the standard deviation.}
    \label{fig:training_performance}
    \vspace{-10pt}
\end{figure}

\begin{itemize}
    \setlength\itemsep{0pt}
    \item[\textbf{Q1:}] Do agents trained on sparse rewards improve over seen data?
    \item[\textbf{Q2:}] How much data is required for task learning?
    \item[\textbf{Q3:}] How important is the quality of the data?
    \item[\textbf{Q4:}] How does multimodality in the data affect training?
    \item[\textbf{Q5:}] Can we use the SMPC costs and data for RL or imitation learning directly?
\end{itemize}

Additional ablations on the percentage of expert data, bounded critic networks, and phase-out thresholds can be found in \autoref{app:ablations}. For more details on our hardware setup, see \autoref{app:deploy}.

\paragraph{Real-world deployment on diverse platforms}
We design five diverse tasks, depicted in \autoref{fig:tasks}, that require complex, whole-body coordination. For Spot, we start with proof-of-concept navigation task. Subsequently, we evaluate tasks requiring sustained physical contact and force exertion: pushing a box, uprighting tire, and rolling a tire. Finally, we evaluate generalization to bipedal systems by testing the G1 humanoid on the box-pushing task. The resulting training runs are depicted in \autoref{fig:training_performance}. Across all tasks, policies achieve near-perfect success rates with little variance between different seeds. The steps required for satisfactory performance grows with task complexity.

Empirically, we find that the policies trained exclusively on sparse rewards can be reliably deployed on the physical hardware across all five tasks and both robotic platforms. \update{We quantify this for Spot across ten runs in \autoref{tab:real_world_success}.}

\begin{table}[ht]
    \centering
    \caption{\update{Real-world success rates for Spot tasks across ten tasks.}}
    \label{tab:real_world_success}
    \setlength{\tabcolsep}{10pt}
    \begin{tabular}{cccc}
    \toprule
    Reach & Box Pushing & Tire Uprighting & Tire Rolling \\ 
    \midrule
    10 / 10 & 9 / 10 & 10 / 10 & 8 / 10 \\
    \bottomrule
    \end{tabular}
\end{table}

This demonstrates that our hierarchical architecture and offline-to-online training pipeline generalize robustly to real-world dynamics without requiring platform-specific reward engineering. Videos of all real-world deployments demonstrating this generalization can be viewed on our project page \url{https://pages.rai-inst.com/smpc2rl/}.

\begin{figure}[ht]
    \centering
    \includegraphics[width=0.99\textwidth]{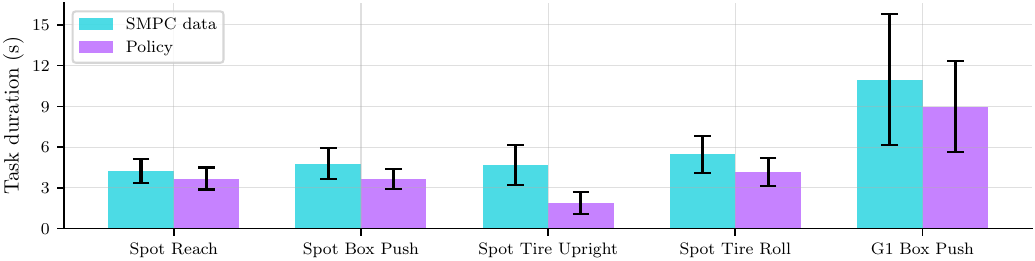}
    \caption{We compare the performance of policies trained on sparse rewards (pink) and the \ac{SMPC} used to generate the offline dataset on all tasks (blue). Our trained policies consistently finish the tasks faster than the \ac{SMPC} experts. \ac{SMPC} statistics are computed over the complete dataset \update{of 4M samples}, policy statistics are averaged over 50k simulated episodes without exploration noise.}
    \vspace{-10pt}
    \label{fig:performance_comparison}
\end{figure}

\paragraph{Q1: Improving over seen expert data}
Sparse rewards eliminate the need for manual tuning and remove the biases introduced by surrogate reward shaping. This enables the discovery of task-optimal behaviors. 
We investigate whether our offline-to-online framework improves over the dataset by comparing the average task-completion time of the trained policy and the \ac{SMPC} expert. As depicted in \autoref{fig:performance_comparison}, our sparse-reward approach consistently outperforms \ac{SMPC}. Some tasks see improvements above 50\%. Notably, learned policies show increased consistency, \update{indicated by a 11-45\% reduction in standard deviation of task duration}. We attribute \ac{SMPC}'s variance to suboptimal trajectories resulting from the limited number of parallel environments used for exploration.

\begin{figure}[ht]
    \centering
    \begin{subfigure}[b]{0.24\textwidth}
        \centering
        \includegraphics[width=\textwidth]{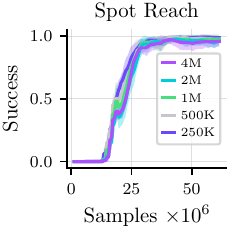}
    \end{subfigure}
    \hfill
    \begin{subfigure}[b]{0.24\textwidth}
        \centering
        \includegraphics[width=\textwidth]{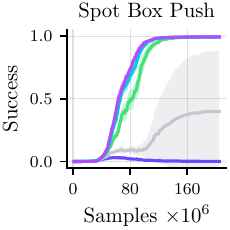}
    \end{subfigure}
    \hfill
    \begin{subfigure}[b]{0.24\textwidth}
        \centering
        \includegraphics[width=\textwidth]{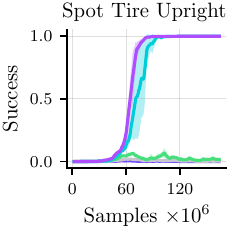}
    \end{subfigure}
    \hfill
    \begin{subfigure}[b]{0.24\textwidth}
        \centering
        \includegraphics[width=\textwidth]{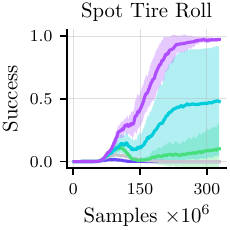}
    \end{subfigure}
    \caption{Training performance with different \ac{SMPC} dataset sizes. For simple tasks (i.e., navigation), reducing the number of samples has little effect. As tasks get more complex, the minimum required dataset size increases. On our most difficult task, agents require four million samples collected within 4 GPU hours to converge.}
    \vspace{-10pt}
    \label{fig:data_per_task}
\end{figure}

\paragraph{Q2: Amount of data}
While our method's data generation can be parallelized across multiple GPUs, a small dataset is generally desirable for efficiency. Therefore, we evaluate the required volume of expert data across different tasks. We ablate the dataset size per task, scaling up from hundreds of thousands to millions of samples. The results in \autoref{fig:data_per_task} demonstrate a clear correlation with task complexity: while easier tasks can be successfully solved with little data, more complex whole-body manipulation tasks require significantly larger datasets. We posit that this effect is caused by the policy gradient estimate. While critics trained on little data yield sufficiently accurate gradient estimates for easy tasks, value estimates for complex tasks with a wider state distribution require training on a dataset with greater coverage.

\begin{figure}
    \centering
    \begin{subfigure}[b]{0.24\textwidth}
        \centering
        \includegraphics[width=\textwidth]{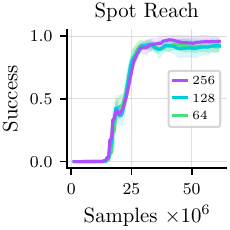}
    \end{subfigure}
    \hfill
    \begin{subfigure}[b]{0.24\textwidth}
        \centering
        \includegraphics[width=\textwidth]{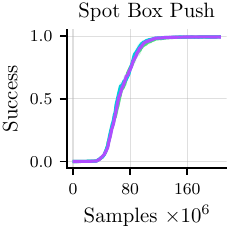}
    \end{subfigure}
    \hfill
    \begin{subfigure}[b]{0.24\textwidth}
        \centering
        \includegraphics[width=\textwidth]{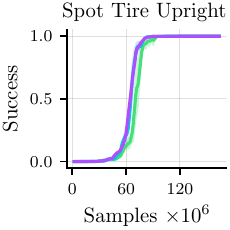}
    \end{subfigure}
    \hfill
    \begin{subfigure}[b]{0.24\textwidth}
        \centering
        \includegraphics[width=\textwidth]{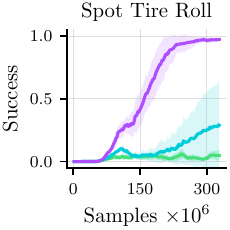}
    \end{subfigure}
    \caption{Performance under varying data quality. We reduce the number of sampling environments per tile during data generation and train policies on the resulting datasets. Most tasks are robust to lower-quality data, whereas tasks that require high coordination are sensitive.}
    \vspace{-10pt}
    \label{fig:trajectory_quality}
\end{figure}

\paragraph{Q3: Quality of the data}
Along the same line, we analyze the influence of the \ac{SMPC} trajectory quality on the downstream \ac{RL} agent. Reducing the number of sampling environments during \ac{SMPC} data generation leads to less exploration of the cost landscape, reducing the likelihood of sampling a \update{well-performing} trajectory (see also \autoref{app:smpc}), and thus resulting in lower-quality expert datasets. Surprisingly, most tasks are relatively robust to a low number of sampling environments (see \autoref{fig:trajectory_quality}). However, for tire rolling, which requires the most coordination, we see a drop in successes. We thus conclude that the required data quality is highly correlated with the required task precision.

\begin{wrapfigure}{r}{0.3\textwidth}
    \centering
    \vspace{-13pt}
    \includegraphics[width=0.3\textwidth]{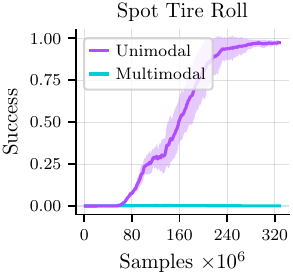}
    \caption{Using multi-modal \ac{SMPC} data heavily affects training and agents fail to learn viable policies.}
    \label{fig:multimodality}
    \vspace{-10pt}
\end{wrapfigure}
\paragraph{Q4: Multimodality in the data}
Our training pipeline collects expert datasets with automated \ac{SMPC}. While \ac{SMPC} is scalable and easily parallelized, it introduces a structural mismatch: \ac{RL} policies rely on the Markov property and output uni-modal action distributions, whereas \ac{SMPC} is inherently multi-modal and not observation-dependent, because it operates directly on the current state and initial guess provided by the \ac{MPC} \update{warm-starting} strategy.

To analyze the impact of multimodality on training stability, we compare training with uni-modal and multi-modal \ac{SMPC} data. Uni-modality is achieved through stricter rewards. Multi-modality appears as, for example, kicking the tire into the goal with Spot's legs, pushing it with the shoulder, or stepping into the tire. Introducing terms for leg and body contacts generates solutions that use the arm instead. Importantly, these terms can be tuned interactively in just a few minutes. As shown in \autoref{fig:multimodality}, the policy trained on the multi-modal dataset completely fails to learn, even though the demonstration success rate is slightly higher, indicating that enforcing a single behavioral mode in the demonstration data is required for successful offline initialization.

\paragraph{Q5: Training directly on the SMPC cost and data}
\update{In the first step of our algorithm, we create an expert policy with \ac{SMPC} with a dense cost function. Another way to exploit the expert is to use its cost function to train RL agents on the dense reward instead of the sparse task reward. Alternatively, we can train on the dataset directly using imitation learning. We investigate this on our Spot tasks using FastTD3~\citep{seo2025fasttd3} for online RL, IQL~\citep{kostrikov2021offlinereinforcementlearningimplicit} for offline RL, and DP~\citep{chi2023diffusionpolicy} for imitation learning.}

\update{The results in \autoref{tab:baselines} show that neither option is performing well. While all baselines converge on the easy navigation task, they fail to reach more than 10\% success rate on more complex tasks. We emphasize that this is not a limitation of the RL algorithms but rather an incompatibility with the \ac{SMPC} cost. Similarly, we attribute the lacking performance of imitation learning to a lack of correction samples in the dataset. DAgger-style setups~\citep{ross2011dagger} that address this limitation are difficult to incorporate because running the \ac{SMPC} alongside RL training is prohibitively expensive.}

\begin{table}
    \centering
    \caption{Comparison against baselines. FastTD3 and IQL use the dense cost function of the \ac{SMPC} expert. IQL and DP train on the SMPC data. We evaluate five runs, each on 100 simulated episodes, with an increased maximum duration to allow slower policies (IQL and DP) to finish.}
    \label{tab:baselines}
    \begin{tabular}{lcccc}
        \toprule
        Task & FastTD3 & IQL & DP & Ours \\
        \midrule
        Reach & \textbf{1.00} & \textbf{1.00} & \textbf{1.00} & \textbf{0.98 $\pm$ 0.03} \\
        Box pushing & 0.00 & 0.03 & 0.10 & \textbf{0.99 $\pm$ 0.005} \\
        Tire uprighting & 0.00 & 0.01 & 0.01 & \textbf{1.00 $\pm$ 0} \\
        Tire rolling & 0.02 & 0.04 & 0.01 & \textbf{0.98 $\pm$ 0.02} \\
        \bottomrule
    \end{tabular}
\end{table}


\section{Conclusion} \label{sec:conclusion}

In this work, we present an offline-to-online \ac{RL} framework that enables the acquisition of complex loco-manipulation skills without manual reward tuning. By leveraging \ac{SMPC} as an automated teacher, we bypass the need for expensive, non-scalable data collection methods such as teleoperation or motion capture. We validate this approach through real-world deployments across five challenging whole-body tasks with an arm-equipped Spot quadruped and a Unitree G1 humanoid.

We investigate the algorithmic conditions required to utilize \ac{SMPC} data for offline-to-online \ac{RL}. By guiding the learning process with expert data, the policy successfully optimizes for sparse task objectives, yielding loco-manipulation behaviors that are quantitatively superior to the dense-reward \ac{SMPC} trajectories. Our ablations show that the required dataset size and quality scale with task complexity, and that filtering the inherent multimodality of the planner is necessary to effectively leverage \ac{SMPC} demonstrations. 

By demonstrating that \ac{SMPC} can be visually tuned in near real-time and massively parallelized for data generation, we position it as a scalable utility for the \ac{RL} community. We hope this work highlights how integrating automated expert data into sparse-reward training can accelerate the development of agile robotic behaviors while avoiding the iteration cycles of manual reward engineering.


\section{Limitations} \label{sec:limitations}


Our offline-to-online framework successfully yields deployable policies, but the achieved behavioral optimality remains local. Because the agent initially relies on \ac{SMPC} demonstrations for successful samples, its policy is tied to the dataset's distribution. Under realistic training constraints, the policy will optimize within this local manifold but is not expected to discover globally optimal strategies that fundamentally diverge from the provided trajectories. Additionally, hyperparameters, such as network sizes and learning rates affect training results.

The current hierarchical control architecture is also limited by the frozen weights of the low-level controller. \update{While it simplifies SMPC and RL training as well as arm-leg coordination, and the generic low-level controller is trained with significant force disturbances on the arms to ensure loads from loco-manipulation are in distribution}, it prevents adaptation to task-specific physical disturbances. This limitation could be resolved by unfreezing weights during late-stage online learning.

Finally, real-world deployment relies on state-based information. Deployment in unstructured environments or outdoors either requires training on vision-based data or a distillation into a vision-based policy.


\clearpage
\acknowledgments{We want to thank Annika Wollschläger and Dario Bellicoso for their help with exporting our policies and deploying them on the G1.}


\bibliography{references}  

\newpage

\appendix

\section{Extended Ablations} \label{app:ablations}

In this section, we perform further ablations on our experiments. All configurations are averaged over five seeds, with the shaded areas denoting the standard deviation.

\begin{figure}[h]
    \centering
    \begin{subfigure}[b]{0.24\textwidth}
        \centering
        \includegraphics[width=\textwidth]{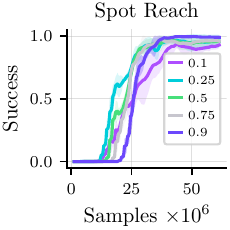}
    \end{subfigure}
    \hfill
    \begin{subfigure}[b]{0.24\textwidth}
        \centering
        \includegraphics[width=\textwidth]{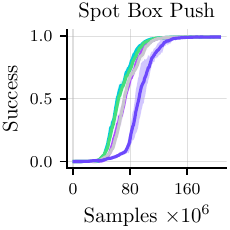}
    \end{subfigure}
    \hfill
    \begin{subfigure}[b]{0.24\textwidth}
        \centering
        \includegraphics[width=\textwidth]{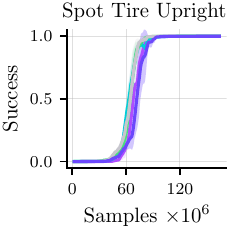}
    \end{subfigure}
    \hfill
    \begin{subfigure}[b]{0.24\textwidth}
        \centering
        \includegraphics[width=\textwidth]{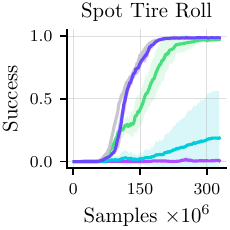}
    \end{subfigure}
    \caption{We ablate the influence of the fraction of expert data in the buffer on training. Most tasks are robust to the choice of the expert data ratio. The exception is tire rolling, which greatly benefits from more expert data initially and collapses below 50\% expert data. Note that all runs additionally use the phase-out threshold of 0.1, so the expert data ratio only directly influences the initial training phase. }
    \label{fig:expert_fraction}
\end{figure}

\paragraph{Percentage of Expert Data} In our experiments, we have fixed the ratio of expert data before phase-outs to 50\%, following the recommendations of \citet{ball2023efficient}. To show that this also holds for data generated by an \ac{SMPC} expert, we run Spot environments with varying ratios of offline to online data. The results are depicted in \autoref{fig:expert_fraction}. 

For reaching, box pushing, and tire uprighting, we find that training runs are almost unaffected by the choice of the expert ratio. This is partly due to our phase-out curriculum, which removes any expert data past a success rate of 10\%. The tire-rolling task, however, is sensitive to the parameter and fails to converge when the amount of expert data is reduced. Notably, the performance of runs with higher expert ratios than 50\% improves more consistently, even after phasing out the expert data. Since the data cannot influence the training after the phase-out point, we posit that exposing the critic to more expert data before improves value estimation around successful states and thereby helps convergence once the policy becomes successful.

\begin{figure}[h]
    \centering
    \includegraphics[width=1.0\textwidth]{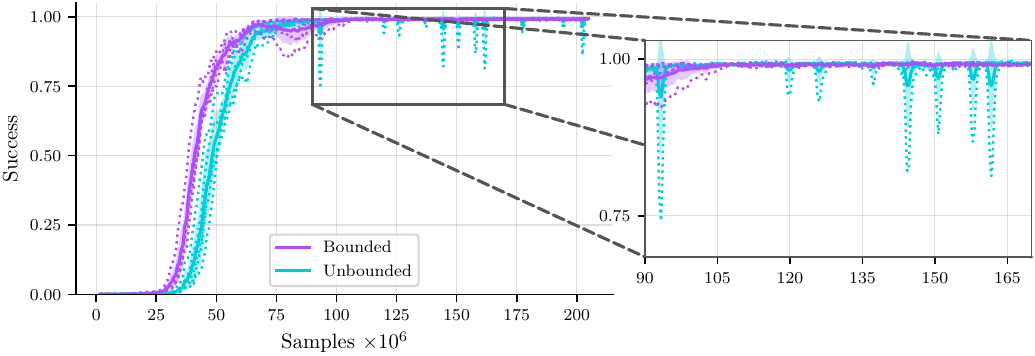}
    \caption{Bounded critics stabilize the training (here Spot box pushing). In parameter ranges at the edges of what is still stable, bounded critics are still well-behaved (pink), whereas unbounded critics start to exhibit sharp drops in performance (blue).}
    \label{fig:bounded_critics}
    \vspace{-10pt}
\end{figure}

\paragraph{Bounded Critic Networks} As outlined in \autoref{sec:method:rl}, we bound the value estimate of the critic network to stabilize the training. While we find that a well-tuned learning rate leads to stable convergence, bounded critic networks increase the range of hyperparameters that still yield successful runs, thereby making tuning easier. We illustrate this by increasing the critic's learning rate to \num{8e-4} and comparing the training curves of unbounded critics with those of our bounded critics on the Spot box pushing task. While both converge, training runs with unbounded critics exhibit spikes where the policy sharply drops in performance and then recovers. To illustrate this better, \autoref{fig:bounded_critics} shows the mean and variance across the two groups, as well as the individual runs as dotted lines.

For more details on how we design the network, see also \autoref{app:subsec:bounded_critic}.

\begin{figure}[h]
    \centering
    \begin{subfigure}[b]{0.24\textwidth}
        \centering
        \includegraphics[width=\textwidth]{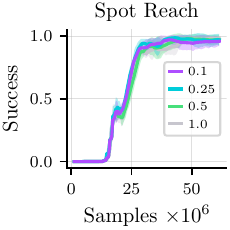}
    \end{subfigure}
    \hfill
    \begin{subfigure}[b]{0.24\textwidth}
        \centering
        \includegraphics[width=\textwidth]{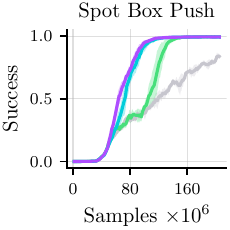}
    \end{subfigure}
    \hfill
    \begin{subfigure}[b]{0.24\textwidth}
        \centering
        \includegraphics[width=\textwidth]{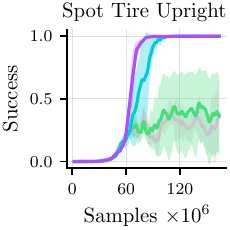}
    \end{subfigure}
    \hfill
    \begin{subfigure}[b]{0.24\textwidth}
        \centering
        \includegraphics[width=\textwidth]{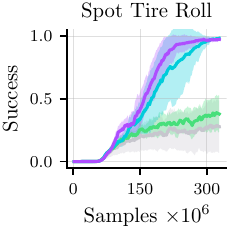}
    \end{subfigure}
    \caption{Ablations on the phase-out threshold. Keeping the offline \ac{SMPC} data in the buffer for too long harms training performance. }
    \label{fig:phase_out}
    \vspace{-10pt}
\end{figure}
\paragraph{Phase-Out Threshold} While the \ac{SMPC} dataset is necessary to bootstrap training on sparse rewards, it is also a suboptimal data source. In \autoref{sec:method:rl}, we therefore introduce a curriculum that phases out the data once runs achieve a 10\% success rate. Here, we examine how different phase-out thresholds affect training across all Spot tasks. The training runs are shown in \autoref{fig:phase_out}.

Surprisingly, we find that keeping the \ac{SMPC} data in the buffer results in significant performance degradation across all tasks except reach. Initially, all runs achieve a success rate of about 10-20\% with the same steep rate of improvement. However, if the data is not removed, training progress slows down noticeably, affecting the thresholds above 25\%. In box pushing, these runs still improve, albeit at a significantly reduced speed. Once the runs with a 50\% required success rate reach their threshold, they resume rapid convergence. Runs that keep the data indefinitely continue to make only slow progress. For tire uprighting and tire rolling, progress slows down so much that runs with a threshold of 50\% or above do not converge within our step limit, although they do improve.

Our explanation for this phenomenon is that while the \ac{SMPC} data is necessary to guide initial exploration, it is too far from the policy to achieve the fine-grained improvements required by the tasks, and it becomes harmful once the policy is narrowed to a successful strategy with a shifted data distribution.

\section{RL Training Details} \label{app:rl}
We train all our policies with \ac{TD3}~\citep{fujimoto2018addressing} and build on the settings proposed by FastTD3~\citep{seo2025fasttd3} to adapt it to training with massively parallel simulation. Our environments are implemented in mjlab with MuJoCo Warp. To ensure robust sim-to-real transfer, we randomize the object mass, its size, and friction. In addition, we use an asymmetric actor-critic setup in which the actor receives noisy observations, while the critic receives exact observations from the simulation. We also use bounded critics, constrained by the theoretical limits of the Q-function. To keep the value predictions close to 0, we scale our rewards. The hyperparameters used across all experiments to train our \ac{RL} policies are given in \autoref{tab:td3_params}. For the G1 task, the difference in joint mappings leads to a slightly increased action regularization of \num{8e-4}, and we increase the critic learning rate to \num{5e-4} to accelerate convergence. \update{Training takes 20 to 90 minutes on an RTX 5090, depending on the complexity of the task.}

Our lower-level policy is trained using the recipe from~\citet{zhu2025versatile} and has been adapted for training in mjlab. We also adapt ReLIC to the G1 for our humanoid experiments. For Spot, we omit torso tilt and height control, as our tasks do not require it. For our G1 experiments, we include it to enable strategies relying on lowering the upper body.

\begin{table}[ht]
    \centering
    \caption{Hyperparameters for the sparse \ac{TD3} training.}
    \label{tab:td3_params}
    \vspace{10pt}
    \setlength{\tabcolsep}{10pt}
    \begin{tabular}{lc}
    \toprule
    Parameter & Value \\ 
    \midrule
    Buffer size $B$ & $16{,}777{,}216 = 2^{24}$   \\
    Gradient steps per update $N_{\nabla}$ & $8$\\
    Phase-out threshold $\eta$ & $0.1$ \\
    Batch size $N_b$ & $8192$   \\
    Discount factor $\gamma$  & $0.99$    \\
    Polyak factor $\tau$ & 0.005 \\
    Actor learning rate $\tau$ & \num{3e-4} \\
    Critic learning rate $\tau$ & \num{1e-4} \\
    Policy delay & 2 \\
    Smoothing noise $\sigma_{\mathrm{s}}$ & 0.05 \\
    Action regularization & \num{5e-4} \\
    Reward scale & 0.01 \\
    \bottomrule
    \end{tabular}
\end{table}

\subsection{Bounding the Critic} \label{app:subsec:bounded_critic}
As previously mentioned, we also bound our critics' Q-value prediction to the theoretical bounds. These are easy to compute because of the simplicity of our reward function. Our best achievable reward is simply 0, and the lowest possible Q-value with $\gamma \in (0, 1)$ is
\begin{align*}
    &\min_{T >= 0} \left(\sum_{i=0}^{T-1} -\gamma ^ i\right) - \gamma ^ {T}\cdot\frac{2}{(1-\gamma)} \\
    &= \left(\sum_{i=0}^{-1} -\gamma ^ i\right) - \gamma ^ {0}\cdot\frac{2}{(1-\gamma)} \\
    &= - \frac{2}{(1-\gamma)}
\end{align*}
taking into account that all robot crash penalties immediately terminate the episode and thus do not accumulate any future rewards. We enforce this by adding a $\tanh$ activation function after the critic's final linear layer and then shifting and scaling it to the desired interval. We also add a constant bias to the prediction before applying $\tanh$, thereby shifting the network output towards zero during initialization. 

\section{SMPC Data Collection} \label{app:smpc}
Our \ac{SMPC} data collection uses the same vectorized mjlab environments we use during training. Because MuJuCo Warp relies on a functional programming paradigm with a single explicit state representation, it lends itself easily to sample-based implementations that need full control over the state. We use this to broadcast states across tiles at the beginning of each episode, and continue to periodically reset all environments to the best candidate chosen for the current batch of samples.

To generate smooth behaviors and prevent high-frequency control noise, the \ac{SMPC} samples actions as splines rather than independent, step-wise commands. As for the \ac{RL} environments, action ranges are normalized to $[-1, 1]$. The planner samples a sparse set of $N_\mathrm{c}$ control points and interpolates them across the planning horizon $T$. After the first iteration, future actions are sampled around the shifted last best action trajectory. To stay reactive, we execute only the first $N_{\mathrm{e}}$ actions, then resample, yielding a \qty{5}{\hertz} policy with \qty{50}{\hertz} action chunks. We discount future rewards in the sampled trajectories with a factor $\gamma$ \update{to bias the objective towards trajectories that achieve success sooner.}

Since data collection computation time is almost exclusively spent running the forward simulation, we save computation by caching the simulation state after $N$ steps and then resetting all simulations to the state of the most successful sample after the full rollout. The high-level pseudocode for our data collection is given in Algorithm~\ref{alg:smpc}. The hyperparameters used for the \ac{SMPC} data collection across all tasks and their meaning are detailed in \autoref{tab:smpc_params}.

\begin{algorithm}
\caption{SMPC Expert Data Collection}
\label{alg:smpc}
\begin{algorithmic}[1]
\State $\mathcal{D} \gets \emptyset$, $\bar{A} \gets 0$ \Comment{Initialize expert dataset $\mathcal{D}$ and warm-start actions $\bar{A}$}
\State $s_{\text{ckpt}} \gets \Call{GetStates}{\text{envs}, M}$ \Comment{$M$ initial tile states}

\While{$|\mathcal{D}| < \text{target}$}
    \State $s_{\text{full}} \gets \Call{Repeat}{s_{\text{ckpt}}, K}$  \Comment{Repeat for $K$ samples}
    \State $\Call{SetStates}{\text{envs}, s_{\text{full}}}$ \Comment{Broadcast $M$ states to $M \times K$ parallel envs}
    \State $A \gets \Call{SampleSplines}{\bar{A}, \sigma_{\mathrm{min}}, \sigma_{\mathrm{max}}, N_c}$ \Comment{Sample action sequences around warm start}
    \State $O,R_{\mathrm{smpc}}, R_{\mathrm{sparse}}, s_{N_e} \gets \Call{SimulateHorizon}{\text{envs}, A}$  \Comment{Batched obs. and rewards}
    \State $R_\tau \gets \sum_{i=1}^T \gamma^i \cdot R_{\mathrm{smpc}} [i]$ \Comment{Compute discounted trajectory reward}
    \State $j \gets \Call{ArgmaxPerTile}{R_\tau}$ \Comment{Find elite trajectory} 
    \State $A^* \gets A[j]$ \Comment{Elite action sequences}
    \State $s_{\text{ckpt}} \gets s_{N_e}[j]$ \Comment{Elite states after $N_e$ steps}
    \State $\bar{A} \gets \Call{ShiftTimeForward}{A^*, N_{e}}$
    \State $\Call{ResetStaleWarmStarts}{\bar{A}}$
    \State $\tau \gets \Call{ExtractCommittedSteps}{O, A, R_{\mathrm{sparse}}, j, N_e}$
    \State $\mathcal{D} \gets \mathcal{D} \cup \tau$
    \If{\textbf{any} episodes finished}
        \State $\Call{DiscardFailedEpisodes}{\mathcal{D}}$
    \EndIf

\EndWhile

\State \Return $\mathcal{D}$
\end{algorithmic}
\end{algorithm}

\begin{table}
    \centering
    \caption{Hyperparameters for the \ac{SMPC} expert.}
    \label{tab:smpc_params}
    \vspace{10pt}
    \setlength{\tabcolsep}{10pt}
    \begin{tabular}{lc}
    \toprule
    Parameter & Value \\ 
    \midrule
    Horizon $T$      & $50$   \\
    Spline control points $N_{\mathrm{c}}$ & $10$\\
    Frequency $f$ & $50$ \\
    Execution steps $N_{\mathrm{e}}$ & $10$   \\
    Discount factor $\gamma$  & $0.99$    \\
    Samples per tile $K$    & $256$   \\
    Minimum noise $\sigma_{\mathrm{min}}$    & $0.5$   \\
    Maximum noise $\sigma_{\mathrm{max}}$    & $1.0$   \\
    \bottomrule
    \end{tabular}
\end{table}

\paragraph{SMPC Tuning in Practice}
\update{We start building the SMPC cost function with a distance-to-goal reward and a robot fall penalty. Then we add additional gripper-to-object rewards, penalties for touching the object with the robot torso or legs, action smoothing, etc. SMPC reward weights are adjusted interactively in our visualizer via sliders in real time. Live tuning requires observing one to three trajectories to observe behavior under different conditions. We tune in one environment, but launch multiple and switch between them to quickly confirm performance under different conditions. Tuning is primarily qualitative, as we want to control the SMPC's overall strategy. We add terms when necessary, e.g., the penalty for contacts between Spot's torso and the tire was added after observing that Spot pushed the tire with its body, a dangerous strategy in the  real world. The multi-modal behavior in our submitted video stems from an SMPC that lacks contact penalties and reflects an intermediate solution observed during tuning. Tuning each task took less than an hour. On average, a live tuning episode requires 10.1s for box pushing, and 14.5s for tire rolling. Observing 100 parameter configurations thus needs $<$30 minutes, whereas RL would take $>$100 hours. The tiled dataset generation for reach, box pushing, tire uprighting, tire rolling and the G1 took 1:39h, 2:41h, 3:28h, 3:15h, and 10:06h, respectively. }

\section{Hardware Setup} \label{app:deploy}
For the quadruped tasks, we deploy the trained policies on a Boston Dynamics quadrupedal Spot robot with an arm. Policy inference is run on an off-board computer for convenience, and commands are sent to the robot via WiFi. The humanoid policies are run directly on a computer mounted on the Unitree G1 robot.

The box is \qty{0.5}{\metre} $\times$ \qty{0.5}{\metre} $\times$ \qty{0.5}{\metre} in size and weighs \qty{1.2}{\kilogram}. The tire has a radius of \qty{0.33}{\metre}, width of \qty{0.34}{\metre}, and weighs \qty{14.3}{\kilogram}.

For state estimation, we use an OptiTrack motion capture system to track the robot and objects in combination with the robots' on-board sensing.

\end{document}